\documentclass[10pt,a4paper,twocolumn]{article}
\usepackage{fontspec}
\usepackage[left=17mm,right=17mm,top=20mm,bottom=20mm,columnsep=7mm]{geometry}
\usepackage{microtype}
\usepackage{booktabs}
\usepackage{array}
\usepackage{amssymb,amsmath}
\usepackage{tikz}
\usetikzlibrary{arrows.meta,positioning,calc}
\usepackage{caption}
\usepackage{titlesec}
\titleformat{\section}{\normalfont\large\bfseries}{\thesection}{0.6em}{}
\titleformat{\subsection}{\normalfont\normalsize\bfseries}{\thesubsection}{0.6em}{}
\titlespacing*{\section}{0pt}{10pt}{4pt}
\titlespacing*{\subsection}{0pt}{8pt}{3pt}
\usepackage[colorlinks=true,linkcolor=black,urlcolor=blue!60!black,citecolor=black]{hyperref}
\hypersetup{pdftitle={Tuning the Stochastic Machine},pdfauthor={George Andrikopoulos}}
\usepackage{url}
\usepackage{balance}
\providecommand{\tightlist}{\setlength{\itemsep}{0pt}\setlength{\parskip}{0pt}}
\newcommand{\FigStack}{%
\begin{figure*}[t]
\centering
\begin{tikzpicture}[
  font=\small,
  lay/.style={draw=black!55, rounded corners=1.5pt, minimum height=8.5mm, text width=52mm, align=center, inner sep=2pt},
  sysl/.style={lay, fill=black!5},
  aill/.style={lay, fill=blue!6},
  lbl/.style={font=\footnotesize\itshape, text=black!60},
  corr/.style={->, >=stealth, draw=black!35, dashed, shorten >=2pt, shorten <=2pt}
]
\node[font=\bfseries] at (0,4.9) {The machine};
\node[font=\bfseries] at (8.2,4.9) {The LLM stack};

\node[sysl] (s1) at (0,4.0) {Silicon / die \\ \scriptsize fixed at the fab};
\node[sysl] (s2) at (0,2.9) {UEFI firmware \\ \scriptsize platform owner, runs first};
\node[sysl] (s3) at (0,1.8) {Loadable kernel modules \\ \scriptsize index resident, body on demand};
\node[sysl] (s4) at (0,0.7) {\texttt{/etc} \\ \scriptsize persistent tunables, read each boot};
\node[sysl] (s5) at (0,-0.4) {RAM \\ \scriptsize cleared at power-off};

\node[aill] (a1) at (8.2,4.0) {Model weights \\ \scriptsize fixed at training};
\node[aill] (a2) at (8.2,2.9) {System prompt \\ \scriptsize platform owner, invisible to user};
\node[aill] (a3) at (8.2,1.8) {Skills / instruction files \\ \scriptsize metadata resident, body on demand};
\node[aill] (a4) at (8.2,0.7) {Rules \& memory files \\ \scriptsize persistent, read each session};
\node[aill] (a5) at (8.2,-0.4) {Context window \\ \scriptsize cleared at session end};

\foreach \i in {1,...,5} { \draw[corr] (s\i.east) -- (a\i.west); }

\draw[->, >=stealth, draw=black!45, line width=0.6pt] (-3.6,-0.9) -- (-3.6,4.4);
\node[lbl, rotate=90, anchor=south] at (-3.85,1.75) {boot order};

\node[align=center, font=\footnotesize, text width=150mm] at (4.1,-1.6)
  {The single operational consequence: a correction made only in the running workload is \texttt{sysctl -w} --- effective now, gone at reboot. Only the persistent layers survive.};
\end{tikzpicture}
\caption{Every session is a cold boot. The context window assembles in layers that correspond, one for one, to a boot sequence; capability written only into the volatile layer does not survive it.}
\label{fig:stack}
\end{figure*}}

\newcommand{\FigLoop}{%
\begin{figure}[t]
\centering
\begin{tikzpicture}[
  font=\footnotesize,
  stp/.style={draw=black!55, rounded corners=2pt, fill=black!4, text width=30mm, align=center, minimum height=7.5mm, inner sep=2.5pt},
  ar/.style={->, >=stealth, draw=black!55}
]
\node[stp] (s1) at (0,4.2)  {\textbf{1} Fix the instance \\ \scriptsize the correction precedes the ceremony};
\node[stp] (s2) at (0,2.9)  {\textbf{2} Name the error class \\ \scriptsize must prevent two \emph{different} future errors};
\node[stp] (s3) at (0,1.6)  {\textbf{3} Draft the rule \\ \scriptsize imperative, with its \emph{why} and a check};
\node[stp] (s4) at (0,0.3)  {\textbf{4} Choose the home \\ \scriptsize global / domain / project --- one home only};
\node[stp] (s5) at (0,-1.0) {\textbf{5} Version and record \\ \scriptsize date, class, incident};

\foreach \a/\b in {s1/s2, s2/s3, s3/s4, s4/s5} { \draw[ar] (\a) -- (\b); }

\node[draw=black!60, fill=blue!7, rounded corners=2pt, text width=26mm, align=center, minimum height=8mm] (art) at (4.6,1.6) {\textbf{Instruction artifact} \\ \scriptsize the persistent layer};
\draw[ar] (s5.east) .. controls (3.4,-1.0) and (4.6,-0.2) .. (art.south);
\draw[ar, dashed] (art.north) .. controls (4.6,3.6) and (3.4,4.2) .. (s1.east)
   node[midway, above, font=\scriptsize, text=black!60, align=center] {loaded at\\next boot};
\end{tikzpicture}
\caption{The write-back path. The loop is closed only when step 5 completes: an artifact that the next cold boot will load. Steps 1--4 without step 5 are a conversation.}
\label{fig:loop}
\end{figure}}

\newcommand{\FigHierarchy}{%
\begin{figure}[t]
\centering
\begin{tikzpicture}[
  font=\footnotesize,
  band/.style={draw=black!55, align=center, minimum height=8mm, inner sep=2pt}
]
\node[band, fill=black!14, text width=62mm] (l1) at (0,3.2) {\textbf{Types} \;\scriptsize deterministic, whole space, compile time};
\node[band, fill=black!10, text width=57mm] (l2) at (0,2.2) {\textbf{Property-based tests} \;\scriptsize deterministic over generated spaces};
\node[band, fill=black!7,  text width=52mm] (l3) at (0,1.2) {\textbf{Unit tests} \;\scriptsize deterministic at sampled points};
\node[band, fill=black!4,  text width=47mm] (l4) at (0,0.2) {\textbf{Constrained decoding} \;\scriptsize where grammar-expressible};
\node[band, fill=blue!7,   text width=42mm] (l5) at (0,-0.8) {\textbf{Natural-language instructions} \;\scriptsize probabilistic residue};

\draw[->, >=stealth, draw=black!55, line width=0.7pt] (4.3,-1.1) -- (4.3,3.5);
\node[rotate=90, anchor=south, font=\scriptsize\itshape, text=black!60] at (4.55,1.2) {push guarantees upward};

\node[font=\scriptsize\itshape, text=black!60, align=left, text width=20mm] at (-4.3,3.2) {binding};
\node[font=\scriptsize\itshape, text=black!60, align=left, text width=20mm] at (-4.3,-0.8) {hints only};
\end{tikzpicture}
\caption{The hierarchy of controls. Prefer the strongest layer that can hold a given guarantee; spend the probabilistic budget only on what cannot be typed or tested. A guarantee held only in prose is unenforced.}
\label{fig:hierarchy}
\end{figure}}

\title{\vspace{-6mm}\textbf{\Large Tuning the Stochastic Machine}\\[1.5mm]
{\large A Systems Engineer's Operating Model for Human-AI Engineering}\vspace{-2mm}}
\author{George Andrikopoulos\\[0.5mm]
{\small Independent researcher, London, United Kingdom}\\[0.3mm]
{\small\url{https://github.com/george-andrikopoulos}}}
\date{\small August 2026}

\begin{document}
\twocolumn[
  \begin{@twocolumnfalse}
  \maketitle
  \begin{abstract}
  \noindent
When an expert corrects an LLM assistant's error, the correction usually dies with the session, and the error class returns. I argue this is an operations problem, not a tooling problem: mechanisms for persisting corrections exist and are shipping, but the \emph{discipline} for governing them --- versioning with provenance, recurrence monitoring, counter-metrics, retirement of stale rules --- does not. Writing as a systems engineer of thirty years, I map the LLM stack onto the machines my profession already operates (frozen silicon, firmware, loadable modules, persistent configuration, volatile memory), identify where the mapping fails (stochastic generation, configuration that binds only probabilistically, no general-purpose retirement (verification) stage by default), and derive from the failures a seven-principle operating discipline with an error loop at its core. Three cases from my own practice illustrate the mechanism, among them a control that silently became the exact harm it was built to prevent. I close with the measurement framework this view implies and the lab study required to test it.

  \end{abstract}
  \vspace{4mm}
  \end{@twocolumnfalse}
]

\hypertarget{the-reboot-problem-capability-is-state-without-a-store}{%
\section{The reboot problem: capability is state without a store}\label{the-reboot-problem-capability-is-state-without-a-store}}

A senior engineer corrects an AI assistant's mistake. The mistake was subtle, plausible, and would have been costly; catching it required years of domain knowledge. The correction is accepted and the work proceeds. A month later, in another session, in a colleague's chat, the same class of mistake appears again, made by the same tool with the same confidence.

Nothing malfunctioned. To see why, one must ask an old systems question of a new system: where does the state live?

Call a component \emph{stateless} when it retains nothing between requests, and \emph{stateful} when it carries memory forward. The distinction is elementary, and it decides everything here, because the modern AI stack is shedding state at every level, deliberately and for good engineering reasons. The model is stateless between sessions: its weights are frozen, and the conversation lives in a context window cleared at session end as thoroughly as RAM at power-off. The plumbing has now followed: the Model Context Protocol, the emerging standard wire between assistants and tools, became stateless at the protocol layer with the specification finalised on 28 July 2026, which removes the initialisation handshake and the session identifier outright, so that any server instance can serve any request with nothing to lose {[}14{]}. The design is correct; statelessness is what one wants from plumbing. I write about that transition as a participant: I have implemented the stateless transport in my own infrastructure, and my open reference design for the connectivity layer above it, Ferridis, is built on the same conviction that evicted state must be explicitly rehoused, not wished away {[}15{]}.

But observe what has happened to the one thing that must not be stateless. The capability of a human-AI engineering pair --- what the pair has learned, which errors the expert has already caught, which corrections have already been made --- is inherently stateful. It is precisely the state that ought to accumulate. And in prevailing practice it has no governed store: it squats in the single place in the stack where destruction is guaranteed, the chat session. The industry names the resulting condition an adoption problem and measures it with usage dashboards --- licences active, prompts issued --- numbers that rise on their own and, by Goodhart's law, become targets rewarding activity over capability {[}1{]}. A team whose AI-assisted output improves every month and a team merely generating more drafts for review are indistinguishable on such a dashboard.

This paper is an experience report and an operating model, not an empirical validation; the study that validation requires is specified in §8. My claim, stated precisely after the related work that bounds it, is that the missing layer is operational discipline, and that the discipline required is one systems engineers already possess.

\hypertarget{related-work-and-the-claim-stated-precisely}{%
\section{Related work, and the claim stated precisely}\label{related-work-and-the-claim-stated-precisely}}

The \emph{mechanisms} for persisting corrections exist, and any honest statement of the problem must begin there. Memory-augmented architectures page long-term state in and out of the context window {[}5{]}; language agents critique and revise their own outputs within and across episodes {[}6{]}{[}7{]}; Voyager maintains an explicit, growing skill library as executable artifacts {[}8{]}; simulated agents accumulate and retrieve memory streams {[}9{]}; and at the training layer, RLHF and Constitutional AI are feedback loops that fold human and principled correction into the weights themselves {[}10{]}{[}11{]}. Closer to daily practice, every major coding assistant now reads persistent instruction files --- repository rules files, user-level configuration, vendor memory features --- and a folk literature of context engineering has grown around them. That practice has itself now been characterised empirically: Jiang and Nam analysed 401 open-source repositories containing Cursor rule files and derived a taxonomy of the project context developers choose to persist, spanning conventions, guidelines, project information, directives addressed to the model, and worked examples {[}17{]}. The operational stakes of these artifacts are no longer speculative: Lulla et al.\ ran coding agents against 124 pull requests across 10 repositories, with and without an \texttt{AGENTS.md} file present, and measured a 28.6\% lower median runtime and 16.6\% lower output-token consumption when one was {[}18{]}. The write-back path, as a mechanism, ships.

What does not ship is the \emph{operations discipline} for these mechanisms at the level of a working team: rules carrying provenance (date, triggering incident, rationale) so they can be audited and retired; recurrence monitoring, so that an error returning against an existing rule is recognised as drift rather than novelty; explicit generalisation from instance to error class, so the library retires classes rather than accumulating anecdotes; counter-metrics against gaming; and a standing cut list, so the library does not decay into sediment. Jiang and Nam's taxonomy {[}17{]} locates the gap precisely: it establishes what practitioners put \emph{into} these artifacts, while the question of how the artifacts are governed over time --- where a rule came from, whether it still fires, when it should be retired --- falls outside its scope, and I have not found it addressed elsewhere. The evaluation literature has lately named the same absence from its own side: Pacchiardi et al.\ argue that assessing a system as a frozen artifact at release structurally ignores the behavioural trajectory of one that keeps learning after deployment --- through in-context learning and memory features already in wide use --- and that evaluation should therefore be centred on trajectories rather than snapshots {[}19{]}. That is the present claim seen from the measurement seat rather than the operator's. The distinction is the one operations management has always drawn between owning tools and running a production system. Karpathy's ``LLM OS'' sketch {[}2{]} and the LLMOps literature occupy adjacent ground, but from the architect's seat: what these systems might become. The present paper is written from the operator's seat --- how to run, govern, and measure the one in front of you, today. My claim is therefore narrow: not that the mechanisms of persistence are lacking --- they are not --- but that persistence without governance does not compound, and that the missing governance layer is nameable, and borrowable wholesale from an older discipline {[}4{]}.

\hypertarget{the-machine-in-the-operators-terms}{%
\section{The machine, in the operator's terms}\label{the-machine-in-the-operators-terms}}

Five axioms describe the stack in the vocabulary of machines I have operated for thirty years.

\textbf{A1 --- The silicon is frozen.} The weights are the die: fixed at the fab (training), unchangeable in use. Model selection is CPU selection, a SKU decision trading capability against cost and latency. What the model natively ``knows'' is data baked into ROM at fab time; the knowledge cutoff is the firmware build date, and anything newer must arrive over the wire, through retrieval or tools, rather than from the stale on-die tables.

\textbf{A2 --- Every session is a cold boot, and RAM is volatile.} The context window assembles in layers, precisely as a boot sequence does. The system prompt is UEFI: platform-owner firmware, first to run, invisible to userland, defining what is possible at all. Skill and instruction files are loadable kernel modules: an index resident at boot, bodies loaded on demand. Rules and memory files are \texttt{/etc}, persistent tunables read at every boot. The conversation is the running workload, in RAM; at session end the RAM clears, completely.

\textbf{A3 --- Natural-language configuration is hints, not law.} I scope this deliberately: where a constraint is grammar-expressible, constrained decoding makes it law --- schema-conformant output can be \emph{enforced}, deterministically, at the sampler {[}12{]}. But the great majority of an engineering team's standards are not grammar-expressible, and for those the comparison with machine configuration breaks down: a \texttt{sysctl} either applies or returns an error, whereas an instruction in context does neither --- it shifts probability mass toward compliance. Honoured with high probability; guaranteed never. Two corollaries matter in practice: conflicting instruction files do not fail loudly the way conflicting configs do --- they silently widen the output distribution; and any guarantee held only in prose must be regarded as unenforced.

\textbf{A4 --- The engine is stochastic} (developed in §4).

\textbf{A5 --- No general-purpose retirement stage ships by default.} (In a processor, the retirement stage is the point at which speculative work is either committed or discarded.) A modern CPU speculates wildly and gets away with it because a downstream checker squashes every mispredicted path before it commits state. The field is building partial analogues --- verifier and critic models, guardrail classifiers, self-consistency sampling {[}6{]}{[}13{]} --- and where they apply they are valuable. But note what the CPU analogy quietly assumes and the LLM lacks: a branch mispredict is \emph{detectable by definition}; the architecture knows ground truth and rolls back. A hallucination is precisely a mispredict for which no in-band detector exists. Whatever verification a given pipeline is to have must therefore be supplied deliberately, from outside: a deterministic checker where one can be built, and a human expert where one cannot.

The mapping is set out in Figure~\ref{fig:stack} and Table~\ref{tab:mapping}.

\FigStack

\begin{table*}[t]
\centering
\caption{The operator's mapping: layers of a machine a systems engineer already runs, and their counterparts in the LLM stack.}
\label{tab:mapping}
\small
\begin{tabular}{@{}p{0.44\textwidth}p{0.44\textwidth}@{}}
\toprule
\textbf{Systems world} & \textbf{LLM world} \\
\midrule
Die / fab & Model weights / training \\
ROM; firmware build date & Parametric knowledge; knowledge cutoff \\
UEFI settings & System prompt \\
Loadable kernel modules & Skills / instruction files \\
\texttt{/etc}, \texttt{sysctl.conf} & Rules files, memory, versioned skills \\
RAM / reboot & Context window / new session \\
\texttt{sysctl\ -w}, runtime only & Chat-only correction \\
Scheduler hints & Natural-language instructions \\
The one true \texttt{sysctl} (grammar/constrained decoding) & Enforceable schema-conformant output \\
Branch mispredict (detectable) & Hallucination (no in-band detector) \\
Retirement / verification stage & Expert review; deterministic checkers \\
Run-to-run variance; jitter & Sampling variance \\
Latency distribution; p99.9 & Output-quality distribution; tail failures \\
\bottomrule
\end{tabular}
\end{table*}

From A2 alone falls the paper's central operational rule, and I state it the way it was first learned, in machine rooms: a correction made only in conversation is \texttt{sysctl\ -w} at runtime --- effective immediately, gone at reboot. \textbf{If it is not written to \texttt{/etc}, it never happened.} Most organisations currently run their most expensive expertise in exactly this mode.

\hypertarget{the-stochastic-engine-and-the-tail}{%
\section{The stochastic engine, and the tail}\label{the-stochastic-engine-and-the-tail}}

At every step the model computes, deterministically, a probability distribution over every possible next token, and then samples from it. The sampling configuration (temperature, nucleus and top-k truncation) shapes how much of the distribution's tail is reachable; and even at temperature zero, batching effects and routing can leave residual run-to-run variance --- jitter is inherent here, not incidental. Output quality is therefore a distribution, never a value.

English usage has drifted into treating ``stochastic'' as a synonym for random, but the word is Greek, and in my mother tongue it never lost its aim. Στόχος: the target. A \emph{stochastikos} was one skilled at aiming under uncertainty --- the archer reading the wind, the physician making the shrewd conjecture. I do not offer the etymology as an argument; I offer it as the correct mental picture, against the incorrect one the modern usage suggests. These are not randomness engines. They are aiming engines: every response a shot at a target with scatter around it, a hallucination the wild shot in the tail, run-to-run variance the jitter of the mechanism.

For an engineer raised on latency work, this picture does most of the necessary conceptual labour by itself. Nobody who has operated a trading system asks what ``the latency'' is; there is only the distribution --- p50, p99, p99.9, max --- and the craft lives in the tail. LLM output is the same object. To evaluate a model on its average response is to evaluate an execution path on mean latency: it hides exactly the spikes that do the damage. A model brilliant at p50 and dangerous at p99 is a system that is fast on average and misses its SLA once a day.

\hypertarget{the-discipline}{%
\section{The discipline}\label{the-discipline}}

Seven principles. I state each with its mechanism; the first four are forced directly by an axiom of §3, the last three by the practical consequences of operating under those four.

\textbf{P1 --- Persist or perish.} Every meaningful correction ends as a durable, versioned artifact in the instruction layer --- the write to \texttt{/etc} --- or it is lost at reboot. The marksman's form: after a miss, adjust the sights, not the archer's memory. The sight adjustment is physical, persistent, and inherited by every future shot, whoever shoulders the rifle.

\textbf{P2 --- Aim by layers; one home per rule.} Instruction artifacts compose in three layers: global (disciplines applying to every task, the error loop among them), domain (rules of a language or field, loaded wherever the domain appears), and project (logic and invariants living with the code, versioned in the repository so they cannot drift from what the project is). To duplicate a domain rule into a project layer is to link a kernel module statically into the binary: the copies fork and rot. When an error recurs against an existing rule, strengthen the original; never lay a duplicate beside it. And conflicting homes are worse than duplicates, for by A3 they do not error --- they widen the distribution.

\textbf{P3 --- Engineer the distribution; judge the tail.} The machine cannot be made deterministic. The work is therefore distribution engineering: move the mean onto the στόχος, pull the variance in, bound the tail; and evaluate on the tail, never the average.

\FigHierarchy
\textbf{P3, corollary --- the hierarchy of controls.} Prefer deterministic controls wherever they exist (Figure~\ref{fig:hierarchy}). In descending strength: types (deterministic over the whole space at compile time; in a pipeline of AI-generated code, a strong type system is a retirement stage in silicon); property-based tests (deterministic over generated input spaces); unit tests (deterministic at sampled points); constrained decoding where the constraint is grammar-expressible; and only then natural-language instructions, the probabilistic residue. Push every guarantee as far up this hierarchy as it will go. This is why type-driven design --- ``parse, don't validate,'' illegal states made unrepresentable {[}3{]} --- is not a stylistic preference in AI-assisted development but a load-bearing safety mechanism, and the cheapest available.

\textbf{P4 --- The human is the retirement stage of last resort.} The expert supplies what A5 says no default pipeline provides: the checker that catches the undetectable mispredict, squashes it, and --- through P1 --- writes the fix back to boot configuration so the pipeline never replays it. This inverts the popular framing, in which the machine displaces the expert. Expertise is not what the machine replaces; expertise is the mechanism that makes the machine safe, and the write-back path is what makes the pair compound rather than merely coexist.

\textbf{P5 --- Every metric carries a counter-metric.} A lone measure defines success incompletely, and rational actors optimise what is defined; the gap between the number and the goal is where the gaming lives {[}1{]}. Speed pairs with rework, conformance with alert noise. Usage, the most gameable measure in the field, is replaced entirely by outcomes (§8).

\textbf{P6 --- Govern or decay.} Context is finite, and the empirical literature confirms what any cache engineer would suspect: retrieval and use of in-context material degrades with length and position {[}16{]}. Every resident rule therefore taxes every other; an ungoverned instruction library does not merely bloat, it degrades its own signal. Hence a review cadence; versioning with provenance --- dated rules carrying their reason and their triggering incident, because rules with provenance can be audited and retired while anonymous rules accumulate as sediment; recurrence monitoring, where an error that returns against an existing rule is drift rather than novelty --- the more important of the two findings; and a standing cut list removing any rule that no longer changes behaviour. This governance layer, not any single mechanism, is what §2 identified as missing from current practice; its industrial ancestor is standard work and continuous improvement in the Toyota tradition {[}4{]}.

Governing the cut list honestly requires confronting what I will call the \emph{deterrence paradox}: a rule that works perfectly leaves no trace, because the error simply never happens --- so the library's best rules score lowest on any naive usage count, exactly as nobody removes a guardrail on the grounds that no one has hit it. Prevention cannot be observed directly; it must be triangulated. Four proxies serve: \emph{deflection citations} (the near-miss report --- the observable moment is not the error but the instant a rule changes a decision), used strictly as a floor signal and itself policed for performative citation, since a signal that feeds a review invites gaming per P5; \emph{trigger-surface measurement}, the actuarial move --- one cannot count fires, but one can count how often the risky context arises, which reframes retirement from the unanswerable ``did it fire?'' to the answerable ``is its road still travelled?''; \emph{recurrence} as the loud negative, outranking any usage statistic; and \emph{challenge trials} --- tasks deliberately constructed to elicit a known error class, run with and without the rule loaded --- the one probe that defeats the paradox outright, because it manufactures the exposure rather than waiting for it. The deeper resolution, though, is lifecycle: the paradox is a property of the instructions layer alone --- mechanised checks do not share it, since a blocked event leaves a log --- so prose is properly a \emph{holding pen}. The healthy trajectory of a rule is incident → prose rule → mechanised check (hook, CI, type) wherever it is checkable --- at which point its measurement problem dissolves, since blocked events leave logs --- and finally an attic, archived outside the loaded paths with provenance intact, never deleted, resurrectable if its class recurs. The cut-list review is the pump that moves rules along this lifecycle; the one verdict it must never issue is ``zero fires, therefore delete.''

\textbf{P7 --- Aim only at real targets.} Before engineering a distribution, verify that a στόχος exists, with an operational test: a target exists where some decision changes on the measured outcome. Some processes are pure scatter --- uniform randomness, noise mistaken for signal, metrics with no decision attached --- and a persuasive framework pointed at a targetless process produces confident, disciplined, worthless work. The first question is never how to aim better; it is whether there is anything to aim at.

\hypertarget{the-loop}{%
\section{The loop}
\FigLoop\label{the-loop}}

The write-back path of P1 and P4 (Figure~\ref{fig:loop}), in the operational form I use daily; readers of the lean tradition will recognise its ancestry {[}4{]}. On every meaningful miss:

\begin{enumerate}
\def\labelenumi{\arabic{enumi}.}
\tightlist
\item
  \textbf{Fix the instance first.} The correction precedes the ceremony; no process may delay the fix.
\item
  \textbf{Name the error class.} Climb the ladder of abstraction until the rule would prevent at least two different future errors, and stop before platitude. Genuinely one-off corrections are exempt: not every correction deserves a rule; every recurring class does.
\item
  \textbf{Draft the rule} as it will stand in the destination artifact: imperative; carrying its reason, since these models generalise from rationales better than from bare commands; and carrying a built-in check wherever the error came from an unverified assumption. The check that has earned its keep in my practice is one question: \emph{under what configuration does this cause the exact harm it was chosen to prevent?} Anchor the rule with the incident, one line, as an example.
\item
  \textbf{Choose the home by layer} (P2), first checking whether a related rule exists. If it does and the error still happened, that is drift; strengthen the rule, do not duplicate it.
\item
  \textbf{Version and record}: date, class, incident, one changelog line.
\end{enumerate}

\hypertarget{evidence-from-practice}{%
\section{Evidence from practice}\label{evidence-from-practice}}

What follows is an experience report: three cases from my own environments, offered as illustrations of mechanism. They establish that the loop is operable and what it feels like in use; they do not and cannot establish effect sizes, which is the work of §8.

\textbf{Case 1 --- a mechanism chosen by reputation.} In a personal low-latency Linux project, an LLM assistant recommended io\_uring for file I/O near isolated CPU cores, on the grounds that it is ``async and low-latency.'' What the model could not know from reputation is that absent \texttt{FMODE\_NOWAIT} support on the target descriptors, submissions fall back to io-wq kernel worker threads, and the scheduler is free to place those threads on the very cores the mechanism was chosen to protect. The choice caused precisely the perturbation it was selected to avoid. The lasting fix was not the code change but the generalised rule --- never select a mechanism by reputation; state what memory, locks, and cores it actually touches; then run the failure-mode question of §6 --- written into the domain instruction layer, dated. The class has not recurred in my use since.

\textbf{Case 2 --- the control that became its own negation.} In my home-lab service mesh, a pre-deployment review of an IP allowlist caught two defects that had survived functional development. First, a configuration handoff produced newline-separated output where JSON was expected: the allowlist would have booted \emph{inert} --- present, passing smoke tests, enforcing nothing. Second, a single trailing character, \texttt{192.168.1.0/} in place of \texttt{192.168.1.0/24}, parsed as \texttt{/0}: match everything. One character separated an allowlist from an allow-everything list. Both defects belong to the class anticipated by the failure-mode question of §6, which Case 1's rule prescribes for design --- a control silently becoming the exact harm it exists to prevent --- and both were caught by asking of the review the same failure-mode question the rule prescribes for design. The generalisation added to the library: a security boundary's parse result must be validated against intent, not merely against syntax; and a guarantee that has never fired is a hypothesis.

\textbf{Case 3 --- the loop running in both directions.} Two episodes from the same period persuade me the discipline is symmetric. In the first, I caught the assistant inserting an estimated biographical fact into a send-ready document; the correction generalised --- never estimate facts about a person in a deliverable; the unconfirmed appears as an explicit placeholder --- and was persisted with provenance. In the second, the gap was mine: a stage of my delivery methodology (an unpublished internal framework) concerned with deployment and adoption had been silently absorbed into adjacent stages rather than named when I restated that methodology as research questions, and it was the assistant's structured review that surfaced it, whereupon the fix was written back into the framework document and versioned. The expert corrects the machine; the machine's review corrects the expert; both corrections persist. I regard this bidirectionality as the strongest qualitative signal in my experience that the unit of improvement is the governed pair, not the model.

\hypertarget{measurement-and-the-study-this-implies}{%
\section{Measurement, and the study this implies}\label{measurement-and-the-study-this-implies}}

If §4's diagnosis is right, valid measurement must be distributional and outcome-based. I propose four instruments, each paired against gaming as P5 demands; I emphasise that these are proposed, and that their validation --- including publication of the severity scale the second instrument requires --- is part of the study below, not an accomplishment of this paper.

\begin{itemize}
\tightlist
\item
  \textbf{Error-class recurrence.} A corrected class must not recur; the series trends to zero as the library matures. Counter: rule usage and a cut list --- zero recurrence is uninformative on its own, since a library padded with dead rules scores the same as one whose live rules are doing the work.
\item
  \textbf{First-time-right rate.} The share of AI-assisted output passing expert review at first pass. Counter: severity-weighted review catches, on a published, versioned severity scale.
\item
  \textbf{Rework rate.} Post-review rework on AI-assisted work; speed that is real, not borrowed. Counter: lead time, so that quality is not faked by slowness.
\item
  \textbf{Time-to-competence.} A new team member to independent delivery, working through the instruction library. Counter: escalation quality, so that juniors still acquire depth rather than outsourcing it.
\end{itemize}

All four roll up into the two currencies any organisation reads --- time saved and risk avoided --- and none can be satisfied by opening more chat windows. Usage measures the activity of aiming; these measure whether the grouping is tightening.

What validation a single-author setting honestly permits is not a field study but a lab study, and I am running it against a versioned reference implementation whose corrections, generalisations, and downstream recurrences are recorded under public provenance. The study's sharpest instrument is the \emph{challenge trial}, as described under P6 in §5: tasks constructed to elicit a rule's error class by a party blind to its exact wording, executed by fresh agents with and without the rule loaded. Challenge trials carry the method because their scoring is deterministic --- no model-in-the-loop grader --- so the with-and-without comparison escapes at once the deterrence paradox and the circularity that dooms self-graded evaluation: the exposure is manufactured, the construction independent, the outcome a binary observable. Around that core sit artifact-level analysis of the instruction library's evolution --- every correction, its generalisation, its recurrence --- and rework and first-time-right rates read against a published, versioned severity scale. Three questions structure the programme. \emph{Mechanism:} which expert corrections generalise into class-retiring artifacts, and what distinguishes them? \emph{Governance:} what operating loop keeps the library effective rather than sedimentary, and do the instruments above validly measure compounding? \emph{Transfer:} developed in one demanding domain, does the discipline move to others, and what predicts the move? A companion paper in preparation takes the challenge trial up directly and reports a first result: one consistently-failing task class closed completely by a single mined rule, beside a suite authored from the rules themselves that found nothing --- locating a discipline's worth in measurement on real misses rather than in its own rulebook. §7's cases and these lab runs are the programme's pilot observations, not its results.

\hypertarget{limits}{%
\section{Limits}\label{limits}}

The mapping is a working analogy, not an isomorphism; context windows are not literally RAM, and the degradation of in-context retrieval is only cache-\emph{like}, though it is at least empirically documented {[}16{]}. The reliability of instruction triggering, on which P1's value partly rests, is underexplored. The evidence here is one practitioner's, from that practitioner's own environments, and I have deliberately reported no quantitative comparisons of my own for that reason. The validation programme of §8 is lab-based by the same honesty: a single-author reference implementation buys clean provenance and deterministic scoring at the cost of ecological validity, and a production baseline, cross-team transfer, and independent expert grading remain the work of a future host --- a site whose design I specify but whose ownership terms I decline to claim before I hold one. And P7 applies to this framework reflexively: whether a given organisation's use of AI has a στόχος worth this machinery is an empirical question, and I decline to assume it.

\hypertarget{conclusion}{%
\section{Conclusion}\label{conclusion}}

Everything proposed here is what marksmen have always done with scattering instruments --- fix the stance, learn the weapon, and after every miss adjust the sights, not the memory --- and what systems engineers have always done with production machines: persist the configuration, prefer the deterministic control, watch the tail, govern the drift. This month the industry finishes evicting state from its plumbing, and rightly; but capability is state, and state must live somewhere, under governance, with provenance, or it does not compound. The machine is new. The discipline is not. It was waiting in the muscle memory of every engineer who ever bounded a jitter distribution; the profession that keeps trading estates in spec already knows how to keep this one.

\begin{center}\rule{0.5\linewidth}{0.5pt}\end{center}

\hypertarget{references}{%
\subsection{References}\label{references}}

{[}1{]} Goodhart, C. (1975). \emph{Problems of Monetary Management: The U.K. Experience.} Papers in Monetary Economics, Reserve Bank of Australia. Popular form due to Strathern, M. (1997).
{[}2{]} Karpathy, A. (2023). \emph{Intro to Large Language Models} (``LLM OS'' framing). Talk, November 2023. https://www.youtube.com/watch?v=zjkBMFhNj\_g
{[}3{]} King, A. (2019). \emph{Parse, Don't Validate.} https://lexi-lambda.github.io/blog/2019/11/05/parse-don-t-validate/
{[}4{]} Ohno, T. (1988). \emph{Toyota Production System: Beyond Large-Scale Production.} Productivity Press.
{[}5{]} Packer, C. et al.~(2023). \emph{MemGPT: Towards LLMs as Operating Systems.} arXiv:2310.08560.
{[}6{]} Shinn, N. et al.~(2023). \emph{Reflexion: Language Agents with Verbal Reinforcement Learning.} arXiv:2303.11366.
{[}7{]} Madaan, A. et al.~(2023). \emph{Self-Refine: Iterative Refinement with Self-Feedback.} arXiv:2303.17651.
{[}8{]} Wang, G. et al.~(2023). \emph{Voyager: An Open-Ended Embodied Agent with Large Language Models.} arXiv:2305.16291.
{[}9{]} Park, J.S. et al.~(2023). \emph{Generative Agents: Interactive Simulacra of Human Behavior.} arXiv:2304.03442.
{[}10{]} Ouyang, L. et al.~(2022). \emph{Training Language Models to Follow Instructions with Human Feedback.} arXiv:2203.02155.
{[}11{]} Bai, Y. et al.~(2022). \emph{Constitutional AI: Harmlessness from AI Feedback.} arXiv:2212.08073.
{[}12{]} Willard, B.T. and Louf, R. (2023). \emph{Efficient Guided Generation for Large Language Models.} arXiv:2307.09702. https://arxiv.org/abs/2307.09702
{[}13{]} Wang, X. et al.~(2022). \emph{Self-Consistency Improves Chain of Thought Reasoning in Language Models.} arXiv:2203.11171.
{[}14{]} Model Context Protocol. \emph{The 2026-07-28 Specification} (final; SEP-2575 removal of the initialization handshake, SEP-2567 removal of the session identifier). https://blog.modelcontextprotocol.io/posts/2026-07-28/ --- released 28 July 2026.
{[}15{]} Andrikopoulos, G. (2026). \emph{Ferridis: an opinionated, type-driven take on agent connectivity, designed to live alongside MCP.} Reference design and Rust implementation; submitted to the Agentic AI Foundation. github.com/george-andrikopoulos/Ferridis
{[}16{]} Liu, N.F. et al.~(2023). \emph{Lost in the Middle: How Language Models Use Long Contexts.} arXiv:2307.03172.
{[}17{]} Jiang, S. and Nam, D. (2026). \emph{Beyond the Prompt: An Empirical Study of Cursor Rules.} arXiv:2512.18925. To appear, International Conference on Mining Software Repositories (MSR) 2026. https://doi.org/10.1145/3793302.3793367

{[}18{]} Lulla, J.L., Mohsenimofidi, S., Galster, M., Zhang, J.M., Baltes, S., and Treude, C. (2026). \emph{On the Impact of AGENTS.md Files on the Efficiency of AI Coding Agents.} arXiv:2601.20404.

{[}19{]} Pacchiardi, L. et al.~(2026). \emph{Continual Learning Requires Evaluating Trajectories.} Preprint. https://doi.org/10.5281/zenodo.20344324

\balance
\end{document}